\documentclass{INTERSPEECH2023}
\usepackage{amsmath,graphicx}
\usepackage{amsfonts,amssymb}
\usepackage{booktabs}

\usepackage{algorithm}
\usepackage{algorithmic}
\usepackage{amssymb}
\usepackage{bbding}
\usepackage{amsmath}
\usepackage{multirow}
\usepackage{enumitem}

\usepackage{newfloat}
\usepackage{listings}

\usepackage{hyperref}
\hypersetup{
    colorlinks=true
    }
    
\newcommand\blfootnote[1]{%
  \begingroup
  \renewcommand\thefootnote{}\footnote{#1}%
  \addtocounter{footnote}{-1}%
  \endgroup
}

\interspeechcameraready

\title{Pushing the Limits of Unsupervised Unit Discovery \\ 
for SSL Speech Representation}

\name{
Ziyang Ma$^1$, 
Zhisheng Zheng$^1$, 
Guanrou Yang$^1$,
Yu Wang$^1$$^2$
Chao Zhang$^3$
Xie Chen$^1$$^\ast$
}
\address{
$^1$Shanghai Jiao Tong University, Shanghai, China  \\
$^2$Shanghai AI Laboratory, Shanghai, China \\
$^3$Tsinghua University, Beijing, China}
\email{\{zym.22, chenxie95\}@sjtu.edu.cn}

\begin{document}

\maketitle

\begin{abstract}
The excellent generalization ability of self-supervised learning (SSL) for speech foundation models has garnered significant attention. 
HuBERT is a successful example that utilizes offline clustering to convert speech features into discrete units for a masked language modeling pretext task. 
However, simply clustering features as targets by k-means does not fully inspire the model's performance. 
In this work, we present an unsupervised method to improve SSL targets. 
Two models are proposed, MonoBERT and PolyBERT, which leverage context-independent and context-dependent phoneme-based units for pre-training. 
Our models outperform other SSL models significantly on the LibriSpeech benchmark without the need for iterative re-clustering and re-training. 
Furthermore, our models equipped with context-dependent units even outperform target-improvement models that use labeled data during pre-training. 
How we progressively improve the unit discovery process is demonstrated through experiments. 
\end{abstract}


\vspace{-0.3cm}
\section{Introduction}
\vspace{-0.1cm}
\label{sec:Introduction}
\blfootnote{Corresponding author$^\ast$. }
Self-supervised learning (SSL) has demonstrated remarkable success in representation learning across various fields, including computer vision~\cite{bao2021beit, he2022masked}, natural language processing~\cite{kenton2019bert, liu2019roberta}, and speech processing~\cite{baevski2020wav2vec, hsu2021hubert}. 
In speech representation learning, SSL methods leverage large amounts of unlabeled audio data in the pre-training stage to construct supervisory signals. 
HuBERT~\cite{hsu2021hubert} as a successful example, thanks to the powerful representation capabilities of Transformer~\cite{vaswani2017attention} architecture and the impressive learning ability of masked language modeling (MLM), exhibits exceptional generalization abilities across various downstream tasks~\cite{yang2021superb, 10096308, TSHUBERT-Zhang2023, chen2022leveraging}. 
To generate pseudo units, HuBERT utilizes an offline $k$-means clustering procedure, which has been shown to be an improvement over previous methods.

Clustering features using $k$-means alone, however, does not fully utilize the model's performance, and iterative re-clustering and re-training entail substantial computational overhead. 
To enhance the quality of SSL targets for pre-training, several concurrent works have been proposed. 
HuBERT-AP~\cite{ren2022speech} proposes generating acoustic pieces as training targets by applying the Byte Pair Encoding (BPE) algorithm to the $k$-means units, leading to performance improvements. 
This method still requires the first iteration of pre-training, which fails to reduce computational costs. 
PBERT~\cite{wang2022supervision} suggests using a pre-trained phoneme recognizer to label units for SSL pre-training. 
CTCBERT~\cite{fan2022ctcbert} proposes incorporating Connectionist Temporal Classification (CTC)~\cite{graves2006connectionist} as an auxiliary loss during pre-training based on PBERT, which further enhances the performance on automatic speech recognition (ASR) tasks. 
However, both methods introduce supervised data during pre-training, which may not be feasible in certain scenarios. 
Additionally, these methods primarily focus on ASR-related tasks, without exploring whether the model is universal after enhancing the target quality. 

The objective of this study is to enhance the representation ability of SSL targets in a fully unsupervised manner during the pre-training stage. 
We first propose MonoBERT, whose targets are generated from a modified version of wav2Vec-U 2.0~\cite{baevski2021unsupervised}. 
This simple improvement has resulted in performance enhancement on ASR tasks compared to the original HuBERT. 
Then, various types of PolyBERT models are investigated, whose targets are context-dependent, and observe continuous performance improvements. 
Finally, we explore whether our approach results in a universally optimized model. 
Our model surpasses the baselines on non-ASR tasks related to the content, semantics, and paralinguistics, demonstrating the high generalization ability of our proposed approach. 

\begin{figure*}[t]
    \centering
    \includegraphics[width=0.8\linewidth]{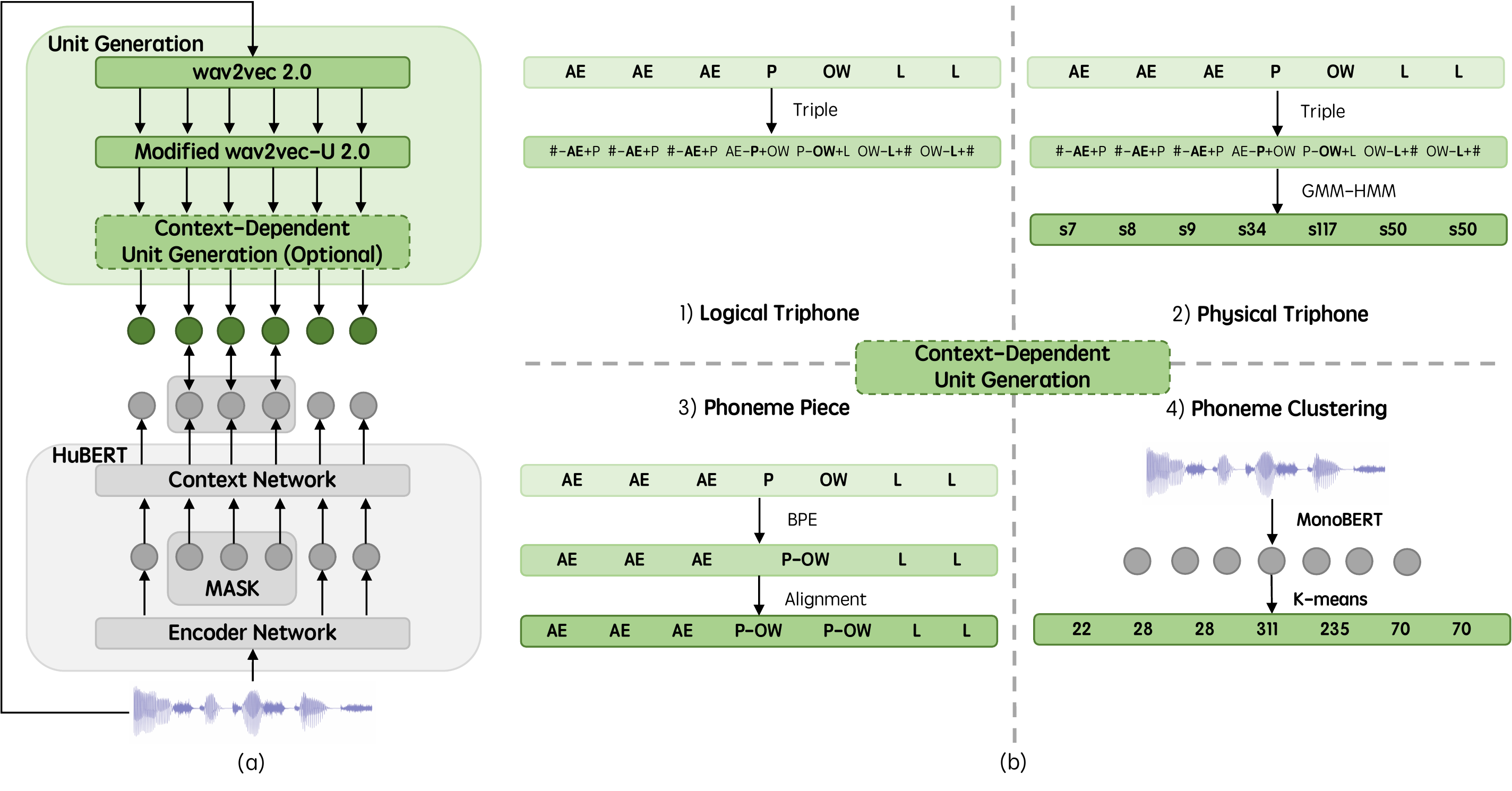}
    \vspace{-0.3cm}
    \caption {(a) illustrates the model structures of MonoBERT and PolyBERT (w/o and w/ Context-Dependent Unit Generation). (b) shows different ways to generate context-dependent units for PolyBERT. All phoneme-based units are obtained with unsupervised methods.}
    \vspace{-0.5cm}
    \label{fig:model}
\end{figure*}

\vspace{-0.3cm}
\section{Prerequisite}
\vspace{-0.1cm}
\label{sec:Prerequisite}
\subsection{Self-Supervised Speech Representation}
SSL pre-training followed by supervised fine-tuning has become a prevalent approach for speech representation learning, with demonstrated efficacy in producing universal representations that can be utilized across various speech downstream tasks~\cite{yang2021superb}. 
This training paradigm typically divides SSL methods into \textbf{1)} contrastive learning~\cite{oord2018representation, schneider2019wav2vec, baevski2019vq, baevski2020wav2vec}, \textbf{2)} predictive learning~\cite{hsu2021hubert, chen2022wavlm, baevski2022data2vec, ma2022mt4ssl, meng2022cobert}, and \textbf{3)} generative learning~\cite{chung2019unsupervised, chung2020vector, liu2021tera, ling2020deep}, based on the pretext tasks employed. 

Predictive learning is the most used SSL method, which predicts pre-clustered or model-generated targets using input features. 
Here we introduce HuBERT, a BERT-like predictive learning SSL model that predicts discrete units of masked regions. 
Given a speech utterance $X=\left[x_1, x_2, \ldots x_L\right]$ of length $L$, HuBERT takes $X$ as the input and outputs a hidden representation $H=\left[h_1, h_2, \ldots h_T\right]$,  where $T$ is the frame number. 
MFCC features extracted from waveform or hidden features derived from a pre-trained model are clustered using the $k$-means algorithm for generating frame-wise discrete units, denoted as $Z=\left[z_1, z_2, \ldots z_T\right]$. 
The model is trained to predict the discrete units $Z$, where each $z_t \in C$ is a $C$-class categorical variable. 
The distribution over the $C$-class units can be written as:
\begin{equation}
p\left(c \mid h_t\right)=\frac{\exp \left(\operatorname{sim}\left(W h_t, e_c\right) / \tau\right)}{\sum_{c^{\prime}=1}^C\exp \left(\operatorname{sim}\left(W h_t, e_{c^{\prime}}\right) / \tau\right)},
\end{equation}
where $W$ is the projection matrix, $e_c$ is the code embedding for unit $c$, $sim$ is the cosine similarity, and $\tau$ is a scaling term. We compute HuBERT loss on the masked $M$ frames, donated as:
\begin{equation}
L=-\sum\nolimits_{t \in M} \log p\left(z_t \mid h_t\right)
\end{equation}

\subsection{Unsupervised Speech Recognition}
Liu et al.~\cite{liu2018completely} demonstrate the first successful unsupervised phoneme recognition through a Generative Adversarial Network (GAN)~\cite{goodfellow2020generative}, which is a proof-of-concept experiment. 
wav2vec-U~\cite{baevski2021unsupervised} is the first model to achieve unsupervised speech recognition on a range of speech recognition benchmarks. 
wav2vec-U 2.0~\cite{liu2022towards} is an optimization of wav2vec-U, reducing the complex and heavy preprocessing steps. 
The core idea of wav2vec-U 2.0 is to train a generator $\mathcal{G}$ and a discriminator $\mathcal{C}$ using adversarial training, which can be written as:
\begin{equation}
\label{eq:wav2vec-U 2.0}
\begin{gathered}
\min _{\mathcal{G}} \max _{\mathcal{C}} \underset{Y_{\mathrm{u}}}{\mathbb{E}}\left[\log \mathcal{C}\left(Y_{\mathrm{u}}\right)\right]+\underset{X}{\mathbb{E}}[\log (1-\mathcal{C}(\mathcal{G}(X)))] \\
-\lambda \mathcal{L}_{g p}+\gamma \mathcal{L}_{s p}+\eta \mathcal{L}_{p d}+\delta \mathcal{L}_{s s},
\end{gathered}
\end{equation}
where $X$ is the sequence of the speech features from a pre-trained wav2vec 2.0, and $\mathcal{G}(X)$ is the transcription output. 
The generator $\mathcal{G}$ is trained to generate phoneme sequences as realistically as possible.
$Y_{\mathrm{u}}$ is the unpaired phoneme sequence with randomly inserted silence. 
The discriminator $\mathcal{C}$ is trained to assign higher scores to real phoneme sequences $Y_{\mathrm{u}}$ and lower scores to generated phoneme sequences $\mathcal{G}(X)$. 
The generator and the discriminator are both parameterized with convolutional neural networks. 
$\mathcal{L}_{g p}$ refers to gradient penalty~\cite{gulrajani2017improved}, $\mathcal{L}_{s p}$ refers to segment smoothness penalty, $\mathcal{L}_{p d}$ refers to phoneme diversity loss, and $\mathcal{L}_{s s}$ refers to auxiliary self-supervised objective. 
$\lambda$, $\gamma$, $\eta$ and $\delta$ are the weights corresponding to the above four losses. 
In the inference stage, only the generator is used to obtain the phoneme-transcription sequences corresponding to speech. 

\vspace{-0.3cm}
\section{Method}
The overall architecture of our models is shown in Figure~\ref{fig:model}~(a), which consists of a unit generation module (Section~\ref{sec:Pseudo Targets Generation}) and a BERT-based backbone network (Section~\ref{sec:Backbone Network}). 
The core idea of our method is to improve the model's representation ability by replacing the SSL targets in the pre-training stage. 
Based on this, we proposed two types of models: 

\textbf{MonoBERT.}
MonoBERT uses the frame-wise monophonic pseudo units generated by the modified wav2vec-U 2.0. 
Instead of conducting the clustering and training loop iteration by iteration as HuBERT, MonoBERT trains the model from scratch for one iteration, without requiring a secondary loop. 

\textbf{PolyBERT.}
PolyBERT uses context-dependent phoneme-based pseudo units generated by the unit generation module. 
The same as MonoBERT, PolyBERT only needs to be trained for one iteration. 
As shown in Figure~\ref{fig:model}~(b), we explore four ways to generate context-dependent units. Details and results are described in Section~\ref{sec:Exploration}. 

\vspace{-0.2cm}
\subsection{Pseudo Units Generation}
\label{sec:Pseudo Targets Generation}
Unlike obtaining the units using the $k$-means clustering algorithm in the original HuBERT paper, we use phoneme-based pseudo units for conducting the following SSL procedure. 
The generation of frame-wise phoneme-based pseudo units is benefited from a modified wav2vec-U 2.0. 
Generally, phoneme sequences generated by the original wav2vec-U 2.0 are not aligned with the speech, which cannot be used as SSL targets. 
Translation invariance is a natural feature of the convolutional neural network which could help. 
After training a convolutional neural network (the generator $\mathcal{G}$), its stride can be adjusted without a constraint. 
When the stride is set to $1$ and deduplication is disabled, monophonic pseudo units are obtained at the frame level, which can be directly utilized for MonoBERT training. 

\vspace{-0.2cm}
\subsection{Backbone Network}
\label{sec:Backbone Network}
To construct the backbone network, we adopt the widely used BERT-based model architecture as mainstream SSL models~\cite{baevski2020wav2vec, hsu2021hubert}, which contains an encoder network and a context network. 
The encoder network is a $7$-layer $1$-D convolutional neural network with kernel sizes of $(5, 2, 2, 2, 2, 2, 2)$ and strides of $(10, 3, 3, 3, 3, 2, 2)$.
It takes the raw audio input $\mathcal{X}$ at a sample rate of $16,000$ Hz and downsamples it to a frequency of $50$ Hz. 
The output representations have a dimension of $512$. 
After that, a linear projection is applied to transform the dimension of representations from $512$ to $768$, followed by a mask matrix to corrupt the representations for conducting the MLM task. 
The context network is a $12$-layer standard Transformer. 
Each Transformer block is set to $768$ model-dimension, $3072$ inner-dimension, and has $8$ attention heads. 
This results in a total of 95M parameters of the model. 
The context network takes the masked version of the $768$-dimension representations as input. 
The final output $\mathcal{H}$ of the backbone is used for computing the HuBERT loss with the prepared offline targets $\mathcal{Z}$. 
The model is optimized to predict discrete units on masked regions. 


\begin{table*}[t]
\begin{center}
\begin{tabular}{l|cc|c|cccc}
\hline
\hline
\multirow{2}*{Model}& \multicolumn{2}{c|}{Pre-training Data} & \multirow{2}*{\shortstack{Language\\Model}}  & \multicolumn{4}{c}{WER$\%$($\downarrow$)}  \\
\cline{2-3}
\cline{5-8}
& Unsupervised & Supervised & & dev-clean & dev-other & test-clean & test-other \\ 
\hline
\hline
\multicolumn{8}{l}{\textit{Self-Supervised Methods}} \\
\hline
\multirow{2}*{HuBERT~\cite{hsu2021hubert}} & \multirow{4}*{960h} & \multirow{4}*{-} & None & 5.5 & 13.0 & 6.3 & 13.2  \\ 
 &  &  & 4-gram & 2.7 & 7.8 & 3.4 & 8.1  \\
\cline{0-0}
\multirow{2}*{WavLM~\cite{chen2022wavlm}} &  &  & None & - & - & 5.7 & 12.0  \\ 
 &  &  & 4-gram & - & - & 3.4 & 7.7  \\ 
\hline
\hline
\multicolumn{8}{l}{\textit{Our Methods}} \\
\hline
\multirow{2}*{MonoBERT} & \multirow{4}*{960h} & \multirow{4}*{-} & None & 4.9 & 11.9 & 5.5 & 11.7  \\ 
 &  &  & 4-gram & 2.7 & 7.3 & 3.2 & 7.6  \\ 
 \cline{0-0}
\multirow{2}*{PolyBERT} &  &  & None & 4.5  & 11.0 & 4.9 & 11.1  \\ 
 &  &  & 4-gram & 2.5 & 7.0 & 3.1 & 7.3  \\ 
\hline
\hline
\multicolumn{8}{l}{\textit{Semi-Supervised Methods}} \\
\hline
\multirow{2}*{PBERT~\cite{wang2022supervision}} & \multirow{4}*{960h} & \multirow{4}*{100h} & None & 4.6 & 11.7 & 4.8 & 11.8  \\ 
 &  &  & 4-gram & 2.6 & 7.3 & 3.2 & 7.7  \\ 
\cline{0-0}
\multirow{2}*{CTCBERT~\cite{fan2022ctcbert}} &  &  & None & 4.6 & 11.3 & 4.8 & 11.3  \\ 
 &  &  & 4-gram & 2.5 & 7.1 &3.1 & 7.4  \\ 
\hline
\hline
\end{tabular}
\caption{WER on LibriSpeech corpus. We compare the performance on four subsets (dev-clean/other \& test-clean/other) with (4-gram) and without (None) language model. The PolyBERT here refers to PolyBERT-PT. The performance of all models comes from their papers except for HuBERT without a language model, for which we fine-tune their public released model. }
\vspace{-1.2cm}
\label{tab:ASR Task Results}
\end{center}
\end{table*}

\vspace{-0.2cm}
\section{Experiments}
\label{sec:Experiments}
\vspace{-0.2cm}

\subsection{Dataset}
\label{sec:Dataset}
In this paper, we conduct our experiments on the mainstream LibriSpeech~\cite{panayotov2015librispeech} 960h-100h benchmark. 
For unsupervised pre-training, we use the full set of the LibriSpeech corpus with $960$-hour (including \texttt{train-clean-100}, \texttt{train-clean-360}, \texttt{train-other-500}) unlabeled data. 
For supervised fine-tuning, the $100$-hour (\texttt{train-clean-100}) split from the LibriSpeech corpus is considered. 
We conduct the evaluation on \texttt{dev-clean/other} and \texttt{test-clean/other} from the LibriSpeech corpus, each of which contains $10$-hour speech data with human-labeled transcriptions. 

\subsection{Implementation Details}
\label{sec:Implementation Details}
To enhance the representation ability of SSL targets during the pre-training phase, our methods maximize the inheritance of the hyperparameters from the HuBERT model without conducting extensive hyperparameter search. 
This demonstrates the effectiveness of our methods and ensures a fair comparison.

\textbf{Pre-Training}.
The pre-training is conducted on NVIDIA GeForce RTX 3090 GPUs from scratch, and we simulate $32$ GPUs by using $8$ GPUs and setting the update frequency to $4$. 
The max token number within a batch is set to $1400000$ on each GPU. 
For the mask strategy, each time-step has a probability of $p = 8\%$ to be selected as the starting index and the subsequent $l = 10$ time-steps are masked. 
For the optimizing strategy, we use Adam~\cite{kingma2015adam} with a peak learning rate of $0.0005$ and a weight decay of $0.01$. 
All models are trained for $400k$ steps, with $8\%$ proportion of warm-up and $92\%$ proportion of linear decay. 

\textbf{Fine-Tuning}. 
The fine-tuning is conducted on $8$ GPUs with a max token number of $3200000$. 
CTC loss is adopted to keep consistent with the baseline models. 
All models are tuned for $80k$ steps, with $[10\%, 40\%, 50\%]$ proportion of warm-up, hold-on, and linearly decay.
For the optimizing strategy, we use Adam with a peak learning rate of $0.00003$. 
During the fine-tuning stage, CNN parameters are fixed permanently and Transformer parameters are fixed for the first $10k$ steps. 
We conduct validation and model selection on the \texttt{dev-other} subset. 

\textbf{Decoding}.
In the decoding stage, we use the Viterbi algorithm or an additional 4-gram language model. The beam size is set to $1500$ during the beam search. 
We conduct testing on \texttt{dev-clean/other} and \texttt{test-clean/other} subsets. 

\subsection{Unit Exploration}
\label{sec:Exploration}
To create MonoBERT, we initially utilize frame-wise context-independent pseudo monophones as SSL targets, as explained in Section~\ref{sec:Pseudo Targets Generation}. 
The vocabulary size in our experiments is $40$, with $39$ monophones and a single silence token. 
As shown in Table~\ref{tab:Exploration}, the MonoBERT yields $10.9/8.5\%$ relative WER reductions on \texttt{dev-clean/other} subsets and $12.7/11.4\%$ relative WER reductions on \texttt{test-clean/other} subsets compared to HuBERT. 
Higher-quality target representations are constructed by fusing contextual information. As shown in Figure~\ref{fig:model}~(b), we explore four ways to conduct the context-dependent unit generation, which make the PolyBERTs:

\textbf{Logical Triphone.}
We generate the logical triphone format of the central monophone from the preceding one and the following one. 
However, this will make the vocabulary size upper bounded by $40^3$, which is too large. 
We select the $500$ most frequently occurring logical triphones in the entire corpus as new tokens, and other logical triphones are still represented by the central monophones, which brings the vocabulary to size $540$. 
We call this model PolyBERT-LT. 
PolyBERT-LT has a similar performance to MonoBERT. 

\textbf{Physical Triphone.}
We build a decision tree to handle the state binding of the sparse logical triphones. 
We use Kaldi\cite{povey2011kaldi} to train a GMM-HMM model and set the leaves number of the tree to $500$, which leads to a number of $448$ states until convergence. 
We use the binded states as the SSL targets and call these units ``physical triphones''. 
We call this model PolyBERT-PT. 
PolyBERT-PT significantly surpasses HuBERT and MonoBERT, proving the effectiveness of the method. 

\textbf{Phoneme Piece.}
We leverage sentencepiece\cite{kudo2018sentencepiece} on the deduplicated pseudo monophones to automatically merge the highly frequent ones into new tokens. 
We call these units ``phoneme pieces''. 
We use open-source code~\footnote{https://github.com/google/sentencepiece} to implement the sentencepiece algorithm and set the vocabulary size to $500$.
Since SSL training needs alignment, we reassign the merged tokens with their corresponding phoneme pieces. 
We call this model PolyBERT-PP. 
PolyBERT-PP also has a significant improvement compared with HuBERT and MonoBERT. 

\textbf{Phoneme Clustering.}
HuBERT adopts iterative re-clustering and re-training to boost the representation ability, which could be helpful to our exploration of target quality. 
Following the unsupervised unit discovery procedure in HuBERT, we use MonoBERT to extract speech features and cluster them with the $k$-means algorithm. 
We set the number of cluster centers to $500$ and take the IDs of the cluster centers as the SSL targets. 
We call this model PolyBERT-PC. 
However, this leads to a performance drop, indicating that phoneme-based units are not suitable for iterative re-clustering. 

\begin{table}[H]
\begin{center}
\resizebox{\linewidth}{!}{
\begin{tabular}{l|cccc}
\hline
\hline
\multirow{2}*{Model} & \multicolumn{4}{c}{WER$\%$($\downarrow$)}  \\
\cline{2-5}
 & dev-clean & dev-other & test-clean & test-other \\ 
\hline
\hline
\multirow{1}*{HuBERT} & 5.5 & 13.0 & 6.3 & 13.2  \\ 
 \hline
\multirow{1}*{MonoBERT} & 4.9 & 11.9 & 5.5 & 11.7  \\ 
\hline
\multirow{1}*{PolyBERT-LT} & 5.1 & 12.0 & 5.3 & 11.7  \\ 
\hline
\multirow{1}*{PolyBERT-PT} & \textbf{4.5}  & \textbf{11.0} & \textbf{4.9} & \textbf{11.1}   \\ 
\hline
\multirow{1}*{PolyBERT-PP} & 4.8 & 11.4 & 4.9 & 11.3  \\ 
\hline
\multirow{1}*{PolyBERT-PC} & 5.3 & 13.9 & 5.6 & 14.3  \\ 
\hline
\hline
\end{tabular}
}
\caption{Exploration towards different context-dependent units. We compare the performance on four subsets (dev-clean/other \& test-clean/other) pre-trained on LibriSpeech train-960 and fine-tuned on train-clean-100 without a language model. }
\vspace{-1cm}
\label{tab:Exploration}
\end{center}
\end{table}

\subsection{ASR Task Results}
\label{sec:ASR Task Results}
Tabel~\ref{tab:ASR Task Results} shows the results of MonoBERT and PolyBERT on the LibriSpeech 960h-100h benchmark compared to other state-of-the-art models. 
The models are pre-trained on LibriSpeech \texttt{train-960}, and fine-tuned on LibriSpeech \texttt{train-clean-100} subset. 
We compare the Word Error Rate (WER) of different models on dev-clean/other and test/other with and without the language model, respectively. 
To sum up, MonoBERT achieves $12.7/11.4\%$ relative WER reduction on \texttt{test-clean/other} subsets over HuBERT without a language model, and $5.9/6.2\%$ relative WER reduction on \texttt{test-clean/other} with a $4$-gram language model. 
While PolyBERT achieves $22.2/15.9\%$ relative WER reduction on \texttt{test-clean/other} subsets over HuBERT without a language model, and $8.8/9.9\%$ relative WER reduction on \texttt{test-clean/other} with a $4$-gram language model. 

Our approach is also compared with other concurrent methods that intend to enhance the quality of SSL targets. 
Despite not using any supervised data during the pre-training phase, PolyBERT surpasses their models, indicating that our techniques produce SSL targets of high quality that are beneficial for the ASR downstream task.

\vspace{-0.3cm}
\subsection{Non-ASR Task Results}
\label{sec:Non-ASR Task Results}
SSL pre-training has been widely proven to be effective on different downstream tasks. 
So it is meaningful to explore whether our context-dependent phoneme-based representations maintain the advantage of SSL pre-training on non-ASR tasks. 
SUPERB~\cite{yang2021superb} is a leaderboard to test the performance of the pre-trained models. 
It provides a series of tasks to investigate four aspects of speech including speaker, content, semantics, and paralinguistics. 
We pick one task from each of the four aspects and evaluate our method: 

\textbf{Speaker Identification (SID, Speaker).}
SID categorizes each utterance into its corresponding speaker identity using a multi-class classification approach. 
The VoxCeleb1 dataset~\cite{nagrani2020voxceleb} is adopted for evaluation. 
The speakers in the predefined set remain consistent across both training and testing. 

\textbf{Keyword Spotting (KS, Content).}
KS identifies preregistered keywords by categorizing utterances into a predefined set of words. 
The Speech Commands dataset v1.0~\cite{warden2017speech} is utilized to perform this task. 
This dataset contains ten classes of keywords, a class for silence, and an additional unknown class to account for false positives. 

\textbf{Intent Classification (IC, Semantics).}
IC categorizes utterances into predefined classes to determine the speaker's intent. 
The Fluent Speech Commands dataset~\cite{lugosch2019speech} is employed for this task. 
Each utterance is associated with three intent labels: action, object, and location in this dataset. 

\textbf{Emotion Recognition (ER, Paralinguistics).}
ER predicts the emotional state of each utterance by assigning it to an emotion class. 
The IEMOCAP dataset~\cite{busso2008iemocap} is utilized. 
The unbalanced emotion classes are excluded and the final four classes are (neutral, happy, sad, and angry). 

Table~\ref{tab:Non-ASR Task Results} shows the results of different pre-trained models on the four non-ASR tasks. 
PolyBERT slightly drops on the SID task. 
A possible explanation is that the phone-based units are independent of the speaker information compared to units generated from a $k$-means model. 
It is amazing that PolyBERT surpasses baselines on KS, IC, and ER tasks, which shows that context-dependent phoneme-based representations have the ability to improve non-ASR tasks including content, semantics, and paralinguistics, demonstrating our methods obtain universal speech representations. 

\vspace{-0.4cm}
\begin{table}[H]
\begin{center}
\begin{tabular}{l|cccc}
\hline
\hline
\multirow{2}*{Model} & \multicolumn{4}{c}{Acc$\%$($\uparrow$)} \\
\cline{2-5}
 & SID & KS & IC & ER \\ 
\hline
\hline
FBANK & 8.5E-4 & 8.63 & 9.10 & 35.39  \\ 
\hline
wav2vec~\cite{schneider2019wav2vec} & 56.56 & 95.59 & 84.92 & 59.79  \\ 
\hline
wav2vec 2.0~\cite{baevski2020wav2vec} & 75.18 & 96.23 & 92.35 & 63.43  \\ 
 \hline
HuBERT~\cite{hsu2021hubert} & \textbf{81.42} & 96.30 & 98.34 & 64.92  \\ 
 \hline
PolyBERT & 77.52 & \textbf{96.66} & \textbf{98.60} & \textbf{65.59}  \\ 
\hline
\hline
\end{tabular}
\caption{Evaluate speech representation ability on various downstream tasks. All four tasks use accuracy (Acc\%) as the evaluation metric. }
\label{tab:Non-ASR Task Results}
\vspace{-1.2cm}
\end{center}
\end{table}
\vspace{-0.6cm}

\section{Conclusion}
\vspace{-0.1cm}
\label{sec:Conclusion}
This paper studies the acquisition of universal speech representations by unsupervised exploration of phone-based SSL units. 
Two models are proposed: MonoBERT and PolyBERT. 
MonoBERT utilizes context-independent phoneme-based units from a modified version of wav2Vec-U 2.0, while PolyBERT employs context-dependent phoneme-based units for SSL pre-training. 
The performance of our models is evaluated on both ASR and non-ASR tasks. 
The results indicate our models perform remarkably better than the baselines in ASR tasks, and even outperform other models that enhance target quality by adding labeled data during pre-training. 
Moreover, other than ASR, our model surpasses the baseline in content-related, semantics-related, and paralinguistics-related tasks. 
These demonstrate that our models produce universal speech representations.

\vspace{-0.5cm}
\section{Acknowledgements}
\vspace{-0.1cm}
This work was supported by the National Natural Science Foundation of China  (No. 62206171 and No.62106140), the International Cooperation Project of PCL, and Alibaba Group through Alibaba Innovative Research Program.

\bibliographystyle{IEEEtran}
\bibliography{mybib}

\begin{thebibliography}{10}
\providecommand{\url}[1]{#1}
\csname url@samestyle\endcsname
\providecommand{\newblock}{\relax}
\providecommand{\bibinfo}[2]{#2}
\providecommand{\BIBentrySTDinterwordspacing}{\spaceskip=0pt\relax}
\providecommand{\BIBentryALTinterwordstretchfactor}{4}
\providecommand{\BIBentryALTinterwordspacing}{\spaceskip=\fontdimen2\font plus
\BIBentryALTinterwordstretchfactor\fontdimen3\font minus
  \fontdimen4\font\relax}
\providecommand{\BIBforeignlanguage}[2]{{%
\expandafter\ifx\csname l@#1\endcsname\relax
\typeout{** WARNING: IEEEtran.bst: No hyphenation pattern has been}%
\typeout{** loaded for the language `#1'. Using the pattern for}%
\typeout{** the default language instead.}%
\else
\language=\csname l@#1\endcsname
\fi
#2}}
\providecommand{\BIBdecl}{\relax}
\BIBdecl

\bibitem{bao2021beit}
H.~Bao, L.~Dong, S.~Piao, and F.~Wei, ``{BEiT: BERT pre-training of image
  Transformers},'' in \emph{Proc. of ICLR}, 2021.

\bibitem{he2022masked}
K.~He, X.~Chen, S.~Xie, Y.~Li, P.~Doll{\'a}r, and R.~Girshick, ``{Masked
  autoencoders are scalable vision learners},'' in \emph{Proc. of CVPR}, 2022.

\bibitem{kenton2019bert}
J.~D. M.-W.~C. Kenton and L.~K. Toutanova, ``{BERT: Pre-training of deep
  bidirectional Transformers for language understanding},'' in \emph{Proc. of
  NAACL}, 2019.

\bibitem{liu2019roberta}
Y.~Liu, M.~Ott, N.~Goyal, J.~Du, M.~Joshi, D.~Chen, O.~Levy, M.~Lewis,
  L.~Zettlemoyer, and V.~Stoyanov, ``{RoBERTa: A robustly optimized BERT
  pretraining approach},'' \emph{arXiv preprint arXiv:1907.11692}, 2019.

\bibitem{baevski2020wav2vec}
A.~Baevski, Y.~Zhou, A.~Mohamed, and M.~Auli, ``{wav2vec 2.0: A framework for
  self-supervised learning of speech representations},'' \emph{Proc. of
  NeurIPS}, 2020.

\bibitem{hsu2021hubert}
W.-N. Hsu, B.~Bolte, Y.-H.~H. Tsai, K.~Lakhotia, R.~Salakhutdinov, and
  A.~Mohamed, ``{HuBERT: Self-supervised speech representation learning by
  masked prediction of hidden units},'' \emph{Trans. of TASLP}, 2021.

\bibitem{vaswani2017attention}
A.~Vaswani, N.~Shazeer, N.~Parmar, J.~Uszkoreit, L.~Jones, A.~N. Gomez,
  {\L}.~Kaiser, and I.~Polosukhin, ``{Attention is all you need},'' \emph{Proc.
  of NeurIPS}, vol.~30, 2017.

\bibitem{yang2021superb}
S.-w. Yang, P.-H. Chi, Y.-S. Chuang, C.-I.~J. Lai, K.~Lakhotia, Y.~Y. Lin,
  A.~T. Liu, J.~Shi, X.~Chang, G.-T. Lin \emph{et~al.}, ``{SUPERB: Speech
  processing Universal PERformance Benchmark},'' \emph{Proc. of InterSpeech},
  2021.

\bibitem{10096308}
W.~Wang and Y.~Qian, ``{HuBERT-AGG: Aggregated Representation Distillation of
  Hidden-Unit Bert for Robust Speech Recognition},'' in \emph{Proc. of ICASSP},
  2023.

\bibitem{TSHUBERT-Zhang2023}
W.~Zhang and Y.~Qian, ``Weakly-supervised speech pre-training: A case study on
  target speech recognition,'' \emph{arXiv preprint arXiv:2305.16286}, 2023.

\bibitem{chen2022leveraging}
C.~Chen, Y.~Hu, Q.~Zhang, H.~Zou, B.~Zhu, and E.~S. Chng, ``Leveraging
  modality-specific representations for audio-visual speech recognition via
  reinforcement learning,'' \emph{Proc. of AAAI}, 2023.

\bibitem{ren2022speech}
S.~Ren, S.~Liu, Y.~Wu, L.~Zhou, and F.~Wei, ``{Speech pre-training with
  acoustic piece},'' \emph{Proc. of Interspeech}, 2022.

\bibitem{wang2022supervision}
C.~Wang, Y.~Wang, Y.~Wu, S.~Chen, J.~Li, S.~Liu, and F.~Wei,
  ``{Supervision-guided codebooks for masked prediction in speech
  pre-training},'' \emph{Proc. of InterSpeech}, 2022.

\bibitem{fan2022ctcbert}
R.~Fan, Y.~Wang, Y.~Gaur, and J.~Li, ``{CTCBERT: Advancing hidden-unit BERT
  with CTC objectives},'' \emph{arXiv preprint arXiv:2210.08603}, 2022.

\bibitem{graves2006connectionist}
A.~Graves, S.~Fern{\'a}ndez, F.~Gomez, and J.~Schmidhuber, ``{Connectionist
  temporal classification: labeling unsegmented sequence data with recurrent
  neural networks},'' in \emph{Proc. of ICML}, 2006.

\bibitem{baevski2021unsupervised}
A.~Baevski, W.-N. Hsu, A.~Conneau, and M.~Auli, ``{Unsupervised speech
  recognition},'' \emph{Proc. of NeurIPS}, 2021.

\bibitem{oord2018representation}
A.~v.~d. Oord, Y.~Li, and O.~Vinyals, ``{Representation learning with
  contrastive predictive coding},'' \emph{arXiv preprint arXiv:1807.03748},
  2018.

\bibitem{schneider2019wav2vec}
S.~Schneider, A.~Baevski, R.~Collobert, and M.~Auli, ``{wav2vec: Unsupervised
  pre-Training for speech recognition},'' \emph{Proc. of Interspeech}, 2019.

\bibitem{baevski2019vq}
A.~Baevski, S.~Schneider, and M.~Auli, ``{vq-wav2vec: Self-supervised learning
  of discrete speech representations},'' in \emph{Proc. of ICLR}, 2019.

\bibitem{chen2022wavlm}
S.~Chen, C.~Wang, Z.~Chen, Y.~Wu, S.~Liu, Z.~Chen, J.~Li, N.~Kanda,
  T.~Yoshioka, X.~Xiao \emph{et~al.}, ``{WavLM: Large-scale self-supervised
  pre-training for full stack speech processing},'' \emph{J. of JSTSP}, 2022.

\bibitem{baevski2022data2vec}
A.~Baevski, W.-N. Hsu, Q.~Xu, A.~Babu, J.~Gu, and M.~Auli, ``{data2vec: A
  general framework for self-supervised learning in speech, vision and
  language},'' \emph{Proc. of ICML}, 2022.

\bibitem{ma2022mt4ssl}
Z.~Ma, Z.~Zheng, C.~Tang, Y.~Wang, and X.~Chen, ``{MT4SSL: Boosting
  Self-Supervised Speech Representation Learning by Integrating Multiple
  Targets},'' \emph{Proc. of InterSpeech}, 2023.

\bibitem{meng2022cobert}
C.~Meng, J.~Ao, T.~Ko, M.~Wang, and H.~Li, ``{CoBERT: Self-Supervised Speech
  Representation Learning Through Code Representation Learning},'' \emph{arXiv
  preprint arXiv:2210.04062}, 2022.

\bibitem{chung2019unsupervised}
Y.-A. Chung, W.-N. Hsu, H.~Tang, and J.~Glass, ``{An unsupervised
  autoregressive model for speech representation learning},'' \emph{Proc. of
  Interspeech}, 2019.

\bibitem{chung2020vector}
Y.-A. Chung, H.~Tang, and J.~Glass, ``{Vector-quantized autoregressive
  predictive coding},'' \emph{Proc. of Interspeech}, 2020.

\bibitem{liu2021tera}
A.~T. Liu, S.-W. Li, and H.-y. Lee, ``{TERA: Self-supervised learning of
  Transformer encoder representation for speech},'' \emph{Trans. of TASLP},
  2021.

\bibitem{ling2020deep}
S.~Ling, Y.~Liu, J.~Salazar, and K.~Kirchhoff, ``{Deep contextualized acoustic
  representations for semi-supervised speech recognition},'' in \emph{Proc. of
  ICASSP}, 2020.

\bibitem{liu2018completely}
D.-R. Liu, K.-Y. Chen, H.-y. Lee, and L.-s. Lee, ``{Completely unsupervised
  phoneme recognition by adversarially learning mapping relationships from
  audio embeddings},'' \emph{Proc. of Interspeech}, 2018.

\bibitem{goodfellow2020generative}
I.~Goodfellow, J.~Pouget-Abadie, M.~Mirza, B.~Xu, D.~Warde-Farley, S.~Ozair,
  A.~Courville, and Y.~Bengio, ``{Generative adversarial networks},''
  \emph{Communications of the ACM}, 2020.

\bibitem{liu2022towards}
A.~H. Liu, W.-N. Hsu, M.~Auli, and A.~Baevski, ``{Towards end-to-end
  unsupervised speech recognition},'' \emph{arXiv preprint arXiv:2204.02492},
  2022.

\bibitem{gulrajani2017improved}
I.~Gulrajani, F.~Ahmed, M.~Arjovsky, V.~Dumoulin, and A.~C. Courville,
  ``{Improved training of Wasserstein GANs},'' \emph{Proc. of NeurIPS}, 2017.

\bibitem{panayotov2015librispeech}
V.~Panayotov, G.~Chen, D.~Povey, and S.~Khudanpur, ``{Librispeech: An ASR
  corpus based on public domain audio books},'' in \emph{Proc. of ICASSP},
  2015.

\bibitem{kingma2015adam}
D.~P. Kingma and J.~Ba, ``{Adam: A method for stochastic optimization},'' in
  \emph{Proc. of ICLR}, 2015.

\bibitem{povey2011kaldi}
D.~Povey, A.~Ghoshal, G.~Boulianne, L.~Burget, O.~Glembek, N.~Goel,
  M.~Hannemann, P.~Motlicek, Y.~Qian, P.~Schwarz \emph{et~al.}, ``{The Kaldi
  speech recognition toolkit},'' in \emph{Proc. of ASRU}, 2011.

\bibitem{kudo2018sentencepiece}
T.~Kudo and J.~Richardson, ``{Sentencepiece: A simple and language independent
  subword tokenizer and detokenizer for neural text processing},'' \emph{Proc.
  of EMNLP}, 2018.

\bibitem{nagrani2020voxceleb}
A.~Nagrani, J.~S. Chung, W.~Xie, and A.~Zisserman, ``{Voxceleb: Large-scale
  speaker verification in the wild},'' \emph{J. of CSL}, 2020.

\bibitem{warden2017speech}
P.~Warden, ``{Speech commands: A public dataset for single-word speech
  recognition},'' \emph{Dataset available online}, 2017.

\bibitem{lugosch2019speech}
L.~Lugosch, M.~Ravanelli, P.~Ignoto, V.~S. Tomar, and Y.~Bengio, ``Speech model
  pre-training for end-to-end spoken language understanding,'' \emph{Proc. of
  InterSpeech}, 2019.

\bibitem{busso2008iemocap}
C.~Busso, M.~Bulut, C.-C. Lee, A.~Kazemzadeh, E.~Mower, S.~Kim, J.~N. Chang,
  S.~Lee, and S.~S. Narayanan, ``{IEMOCAP: Interactive emotional dyadic motion
  capture database},'' \emph{Language resources and evaluation}, 2008.

\end{thebibliography}

\end{document}